\documentclass[10pt,twocolumn,letterpaper]{article}

\usepackage{cvpr}              

\usepackage{xcolor}
\usepackage{algorithm}
\usepackage{algorithmic}
\usepackage{amsmath}
\usepackage{amssymb}

\usepackage{booktabs}
\usepackage{multirow} 
\usepackage{caption}
\usepackage{subcaption}
\usepackage{makecell}
\usepackage{tabularx}
\usepackage{enumitem}

\definecolor{cvprblue}{rgb}{0.21,0.49,0.74}
\usepackage[pagebackref,breaklinks,colorlinks,allcolors=cvprblue]{hyperref}

\def\paperID{} 
\def\confName{CVPR}
\def\confYear{2026}

\title{Region-Wise Correspondence Prediction between Manga Line Art Images}

\author{Yingxuan Li\textsuperscript{\rm 1, 2}, 
Jiafeng Mao\textsuperscript{\rm 2}, 
Qianru Qiu\textsuperscript{\rm 2}\thanks{Corresponding Author}, 
Yusuke Matsui\textsuperscript{\rm 1}\\
\textsuperscript{\rm 1}The University of Tokyo, 
\textsuperscript{\rm 2}CyberAgent, Inc.\\
\texttt{\small \{li\_yingxuan, jiafeng\_mao, qiu\_qianru\}@cyberagent.co.jp, matsui@hal.t.u-tokyo.ac.jp}\\
}

\begin{document}
\maketitle
\begin{abstract}
Understanding region-wise correspondences between manga line art images is fundamental for high-level manga processing, supporting downstream tasks such as line art colorization and in-between frame generation. 
Unlike natural images that contain rich visual cues, manga line art consists only of sparse black-and-white strokes, making it challenging to determine which regions correspond across images. 
In this work, we introduce a new task: \textbf{\textit{predicting region-wise correspondence between raw manga line art images without any annotations}}. 
To address this problem, we propose a Transformer-based framework trained on large-scale, automatically generated region correspondences. 
The model learns to suppress noisy matches and strengthen consistent structural relationships, resulting in robust patch-level feature alignment within and across images. 
During inference, our method segments each line art and establishes coherent region-level correspondences through edge-aware clustering and region matching.
We construct manually annotated benchmarks for evaluation, and experiments across multiple datasets demonstrate both high patch-level accuracy and strong region-level correspondence performance\footnote{The dataset and code are available at: \url{https://github.com/liyingxuan1012/r2r-lineart-correspondence}.}, achieving 78.4--84.4\% region-level accuracy. 
These results highlight the potential of our method for real-world manga and animation applications.
\end{abstract}

\section{Introduction}
\label{sec:intro}

\begin{figure}[t]
\centering
    \includegraphics[width=\linewidth]{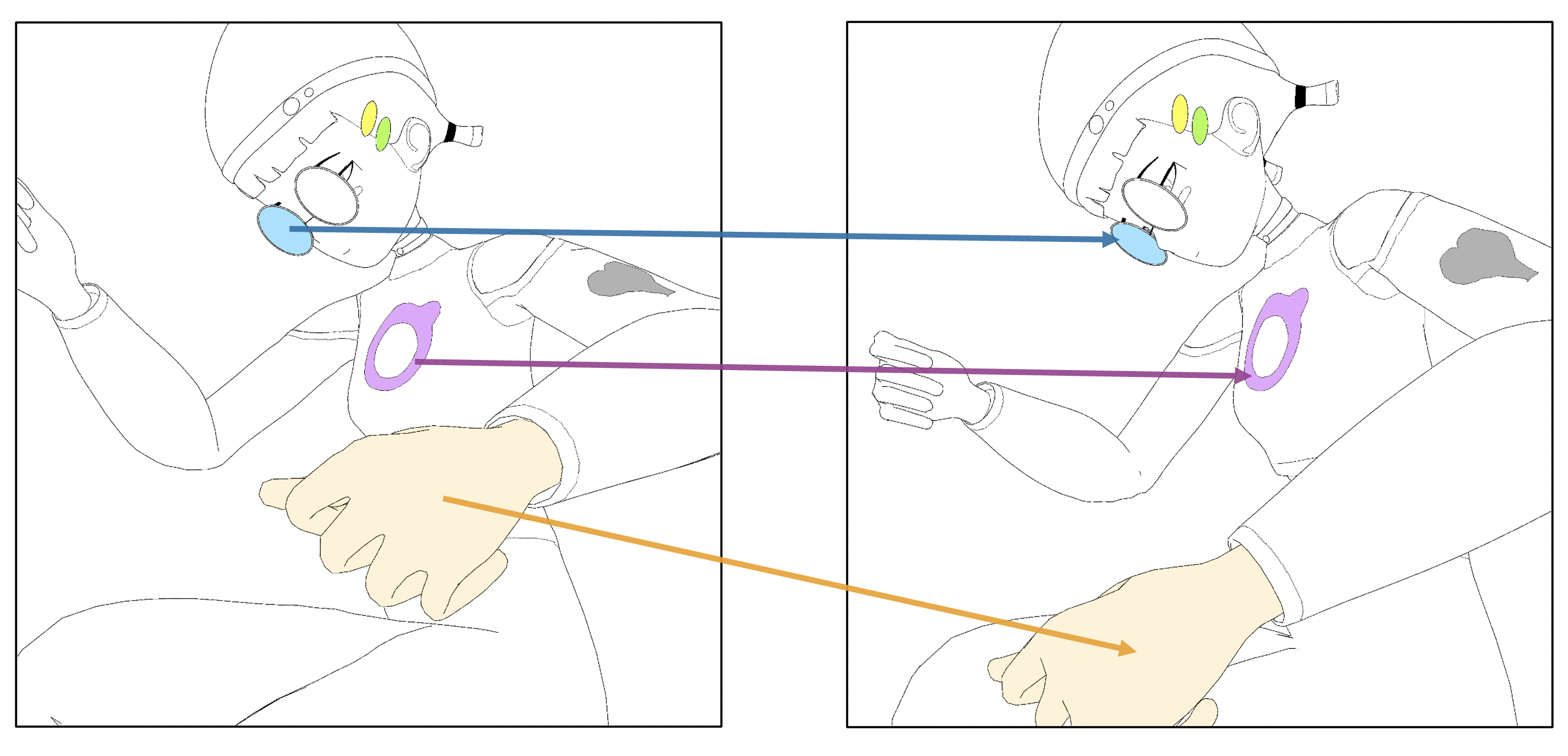}
    \caption{Given a pair of raw manga line art images, our task is to identify meaningful structural regions and match the corresponding regions across the two images. Regions shown with the same color indicate predicted correspondences.}
  \label{fig:intro}
\end{figure}

In traditional manga and animation production, artists often need to manually identify and track semantic regions across numerous line art frames to maintain consistent character appearance. 
For example, a character’s hair must remain the same shade of blue across scenes, and accessories such as a right-ear earring must persist in both shape and position. 
This manual process is highly labor-intensive and requires considerable precision and artistic expertise.  
To address this issue, we explore the fundamental problem of understanding structural regions in line art and predicting region-wise correspondences between different line art images, as shown in Figure~\ref{fig:intro}.  
Such fine-grained structural understanding enables a wide range of downstream applications, including automatic line art colorization and generating visually consistent in-between frames from keyframes.

However, predicting region-wise correspondence in line art is particularly challenging. 
Unlike natural images, line art images are abstract and sparse, lacking texture and color cues that usually guide visual matching.  
Characters may appear in different poses, scales, or viewpoints, and the degree of abstraction varies across artists’ drawing styles, further complicating correspondence estimation.
A more fundamental challenge is the absence of datasets with region-level segmentation annotations. Existing segmentation models are mostly trained on natural images, whose regions have clear and well-defined boundaries.
In contrast, regions in line art are defined by hand-drawn contours that are often open or loosely formed, making them fundamentally different from natural-image regions and difficult to segment accurately.  
Due to both the abstract nature of line art and the lack of effective segmentation methods, region-wise correspondence prediction remains largely underexplored.

Previous works on correspondence prediction in line art often assume that the images have already been segmented into closed regions~\cite{casey2021animation, dai2024learning, feng2025unified}. However, this assumption rarely holds in practical scenarios.  
In this paper, we address a new task: \textbf{\textit{predicting region-wise correspondence directly from raw line art images without any prior segmentation or annotations}}.  
Given a pair of unannotated line art images, our goal is to identify semantically meaningful regions in each image and predict the correspondences between them.

Considering the abstract nature of line art and the challenges of this task, we propose a Transformer-based pipeline that jointly learns intra-image and cross-image features.
We first divide each raw line art image into patches and feed them into a composite Transformer model, which extracts patch-level features and computes patch-wise similarities. 
We then merge adjacent patches with high intra-image similarity to form initial region groups.
Because these patch-based groups produce blocky boundaries, we introduce an edge-aware refinement step to recover pixel-level regions with smooth, stroke-aligned contours, effectively addressing the difficulty of region extraction in line art.  
Finally, we aggregate cross-image patch similarities to produce coherent region-wise correspondence predictions.

To support training without human annotations, we develop an automatic pipeline for pixel-level region segmentation and region-wise matching, producing a large-scale pseudo-labeled dataset.  
For evaluation, we manually refine a subset of these pseudo labels to construct high-quality benchmark datasets.  
We conduct experiments on multiple manga line art datasets, including both synthetic and real-world examples.  
Experimental results demonstrate that our method accurately captures patch-level similarities and produces coherent region-level correspondences across images.

To summarize, our contributions are as follows:
\begin{itemize}
    \item We are the first to study region-wise correspondence prediction directly from raw manga line art images, enabling deeper structural understanding of line art.
    \item We propose a Transformer-based framework that learns patch-level similarities and produces coherent region-level correspondences, demonstrating its effectiveness across diverse datasets.
    \item We develop an automatic annotation pipeline and construct a manually refined benchmark dataset with region-level ground-truth correspondences to support quantitative evaluation.
\end{itemize}
\section{Related Work}
\label{sec:related_work}

\subsection{Manga Line Art Datasets}



Manga line art datasets can be broadly categorized by how the line art is obtained, which largely determines their structural complexity and realism.

A realistic setting extracts line art from real anime frames using learned detectors such as LineArtDetector~\cite{zhang2023adding} in MangaNinja~\cite{liu2025manganinja}. 
The resulting line art exhibits open contours, abstract shapes, and diverse styles, making correspondence prediction challenging.  
ATD-12K~\cite{siyao2021deep} is a representative dataset containing 12,000 keyframe pairs from professional animation, which we convert to line art and evaluate our method under these less structured conditions.

Another line of work derives contours from 3D-rendered characters, producing clean, closed regions with well-defined semantics that are easier to segment and annotate.  
Examples include AnimeRun~\cite{siyao2022animerun} and PaintBucket-Character (PBC)~\cite{dai2024learning}, the latter providing over 10,000 images with part-level labels.

\subsection{Matching in Line Art}
Recent studies in line art processing have explored point-to-point matching for tasks such as automatic colorization~\cite{meng2025anidoc, liu2025manganinja}. 
These methods rely on techniques developed for natural image matching~\cite{lindenberger2023lightglue}, which make use of visual cues such as color and texture. 
However, such cues are absent in line art, leading to sparse and unreliable keypoints.



More recent work has explored region-level matching for animation and colorization~\cite{casey2021animation, dai2024learning, feng2025unified}.
These methods either assume pre-segmented regions or rely on a two-stage pipeline, where segmentation is made feasible by the presence of clean, closed contours that allow reliable rule-based extraction.
However, this assumption does not hold for raw, hand-drawn manga line art, which typically contains open and ambiguous structures.
In contrast, our method does not rely on predefined regions or explicit segmentation, but instead jointly models intra-image structure and inter-image correspondence.
\section{Approach}

\begin{figure*}[t]
    \centering
    \includegraphics[width=\linewidth]{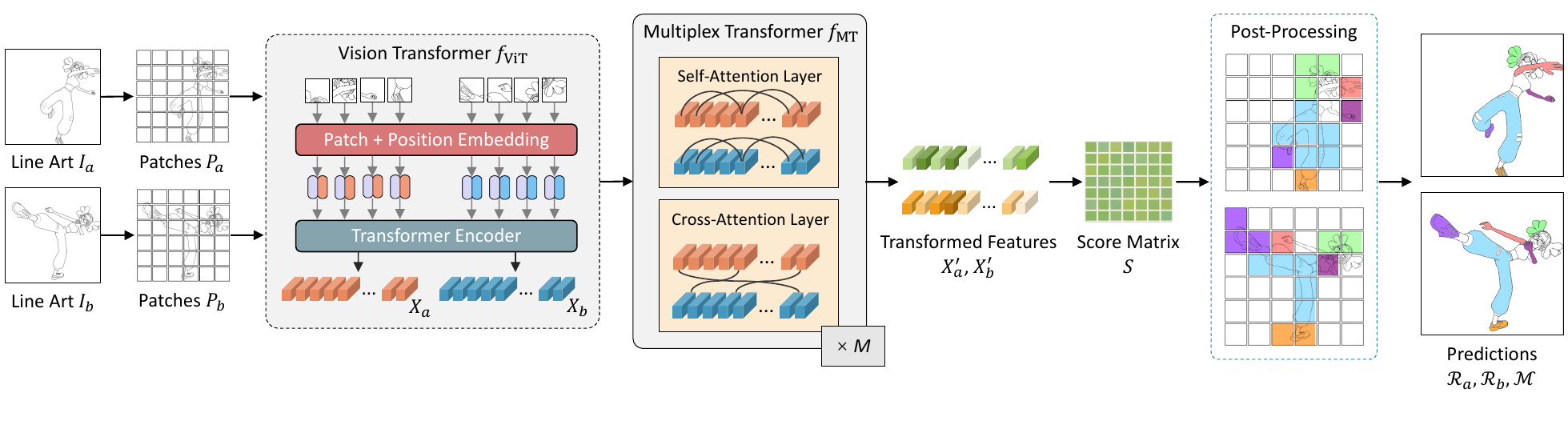}
    \caption{Overview of our proposed pipeline for predicting region-wise correspondence between manga line art images. The model extracts patch-level features using a Vision Transformer and predicts patch-level similarity via a Multiplex Transformer, followed by post-processing to obtain region-level correspondence.}
    \label{fig:pipeline}
\end{figure*}

In this section, we introduce our proposed framework, which consists of two main components: \textbf{\textit{a Transformer-based model for patch-level similarity learning}}, and \textbf{\textit{a post-processing method to predict region-wise correspondences}}.
Figure~\ref{fig:pipeline} illustrates how our method takes a raw pair of manga line art images as input and produces region-wise correspondence predictions.

Given an unannotated pair of line art images $I_a$ and $I_b$, we first divide each image into $N$ patches of size $p \times p$, denoted as $P_a, P_b \in \mathbb{R}^{N \times p \times p}$. These patches are embedded and processed by a Vision Transformer $f_{\text{ViT}}$ to extract patch-level features.
We then use a Multiplex Transformer to model intra-image structures and cross-image relationships, and compute a dense patch similarity matrix $S \in [0,1]^{2N \times 2N}$.
Finally, we apply a post-processing step that first clusters similar patches into region groups $\mathcal{R}_a$ and $\mathcal{R}_b$ based on the intra-image similarities in $S$, and then predicts region-level correspondences $\mathcal{M} \subseteq \mathcal{R}_a \times \mathcal{R}_b$ based on the cross-image similarities in $S$.

In the following sections, we describe each component of our pipeline in detail.

\subsection{Patch-Level Similarity Learning}
\label{sec:approach_patch}

\noindent\textbf{\textit{Feature Extraction.}}  
Different from conventional matching approaches (e.g., point-to-point matching~\cite{sun2021loftr, wang2024efficient, xue2025matcha}) that often rely on CNN-based feature extractors, we adopt a Vision Transformer (ViT)~\cite{dosovitskiy2020image} to learn patch-level features, which are better suited for capturing region-level semantic structures in line art images.

Given patch sets $P_a$ and $P_b$ extracted from the two images, we first map each patch into a $d$-dimensional token via a linear embedding layer, yielding $Z_a, Z_b \in \mathbb{R}^{N \times d}$.  
We then add learnable positional embeddings $E_{\text{pos}} \in \mathbb{R}^{N \times d}$ to encode spatial layout:
\begin{equation}
    \tilde{X}_a = Z_a + E_{\text{pos}}, \quad 
    \tilde{X}_b = Z_b + E_{\text{pos}}.
\end{equation}
The Vision Transformer encoder processes the resulting sequences through multiple stacked self-attention layers to produce patch-level feature representations:
\begin{equation}
    X_a = f_{\text{ViT}}(\tilde{X}_a), \quad 
    X_b = f_{\text{ViT}}(\tilde{X}_b),
\end{equation}
where $X_a, X_b \in \mathbb{R}^{N \times d}$ denote the final patch-level features of images $I_a$ and $I_b$.

\noindent\textbf{\textit{Multiplex Transformer.}}  
We feed the extracted features $X_a, X_b$ into the Multiplex Transformer $f_{\text{MT}}$, which consists of $M$ stacked layers, each containing both self-attention and cross-attention blocks~\cite{vaswani2017attention}.  
At each layer, tokens from both $X_a$ and $X_b$ perform self-attention and attend to tokens in the other image via cross-attention. This design enables the model to capture both intra-image structure and cross-image correspondence simultaneously.

After $M$ blocks, we obtain the transformed patch features $X'_a, X'_b \in \mathbb{R}^{N \times d}$:
\begin{equation}
    X'_a, X'_b = f_{\text{MT}}(X_a, X_b).
\end{equation}
These transformed features are used in the next stage to compute patch-wise similarity across the two images.

\noindent\textbf{\textit{Similarity Computation.}}  
Given the transformed patch features $X'_a, X'_b$, we compute a unified similarity matrix $S \in [0, 1]^{2N\times 2N}$, which contains both intra-image and cross-image similarities, using cosine similarity followed by row-wise softmax.
The matrix $S$ encodes intra-image similarity (within $I_a$ and $I_b$) as well as cross-image similarity (between $I_a$ and $I_b$) in a unified structure.

\noindent\textbf{\textit{Ground-Truth Matrix.}}  
We construct the ground-truth patch correspondence matrix $G \in \{0,1\}^{2N \times 2N}$ as follows.
After dividing each image into $N$ patches, we assign each patch a region ID based on the dominant region contained in it. Specifically, we identify the region that covers the largest area within the patch and exceeds 55\% of the patch area, and use its ID as the patch’s ID.
Two patches within the same image are considered to belong to the same region if they share the same region ID.
For cross-image correspondences, we mark entries in $G$ as 1 for all patch pairs whose underlying regions are annotated as corresponding across the two images ($I_a$ and $I_b$).

\noindent\textbf{\textit{Loss Function.}}  
Since the ground-truth matrix $G$ is highly sparse, naively supervising all $2N \times 2N$ pairs makes training hard to converge and limits performance. To cope with this, we adopt a CLIP-style sampled contrastive loss~\cite{radford2021learning}. 
During training, we randomly draw a subset of positive pairs from the nonzero entries of $G$. For each sampled positive $(i, j)$, we choose $K$ negatives $\{j_k\}_{k=1}^K$ that are not matched with $i$. We then compute a temperature-scaled softmax over $\{j\}\cup\{j_k\}_{k=1}^K$ in the $i$-th row of $S$ and minimize the negative log-likelihood:
\begin{equation}
    \mathcal{L}_{i,j} = - \log \frac{\exp(S_{ij} / \tau)}{\exp(S_{ij} / \tau) + \sum_{k=1}^{K} \exp(S_{ij_k} / \tau)},
\end{equation}
where $\tau$ is a temperature parameter. The final loss is averaged over all sampled positives in the batch, enabling the model to learn meaningful patch-level similarities under sparse supervision.

\subsection{Region Matching via Post-Processing}
\label{sec:approach_region}

\noindent\textbf{\textit{Intra-Image Patch Merging.}}
To derive region-level correspondences from the patch-level similarity matrix produced by our Transformer-based model, we first perform intra-image patch merging as a post-processing step.  

\begin{figure}[t]
\centering
    \includegraphics[width=\linewidth]{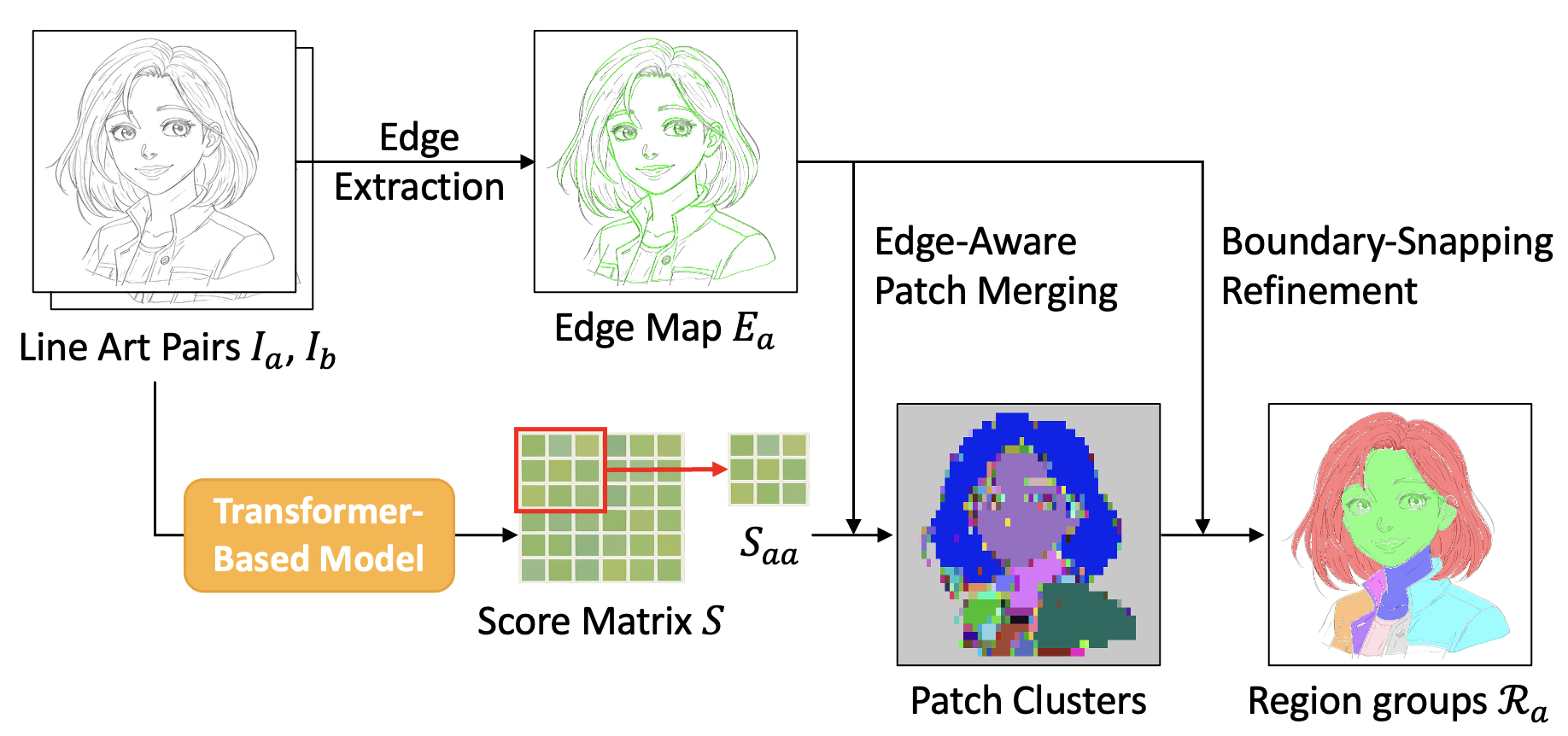}
    \caption{Framework of intra-image patch merging. From the line-art edge map and patch clusters, our edge-aware merging with watershed refinement produces a set of pixel-level, edge-aligned region groups.}
  \label{fig:intra_merge}
\end{figure}

As illustrated in Figure~\ref{fig:intra_merge}, for image $I_a$, we extract the intra-image similarity submatrix $S_{aa} \in [0,1]^{N \times N}$ from the upper-left block of the full similarity matrix $S$, which represents pairwise similarities among all patches within $I_a$ (similarly, $S_{bb}$ is used for $I_b$).
We compute a structural edge map $E_a$ from $I_a$ by applying Gaussian smoothing followed by Sobel gradients~\cite{sobel1968}.
Adjacent patches within an 8-neighborhood are merged based on their similarity in $S_{aa}$, while merges across strong edges are suppressed by checking the average edge response along shared boundaries.  
We then apply a watershed algorithm on $E_a$ using the merged patch clusters as seeds, so that the resulting region boundaries align with the stroke structure.  
To avoid fragmented segmentation, we further merge small regions based on contact length and edge strength in an edge-aware manner.  
Finally, each patch is assigned a region ID via majority voting over its pixels, producing the final pixel-wise region map $\mathcal{R}_a$.

\noindent\textbf{\textit{Cross-Image Region Matching.}}  
Given the region sets $\mathcal{R}_a$ and $\mathcal{R}_b$, we derive region-level similarities by aggregating the cross-image patch similarity submatrices $S_{ab}, S_{ba} \in [0,1]^{N \times N}$.  
Here, $S_{ab}$ corresponds to the upper-right block of the full similarity matrix $S$ and encodes patch-wise similarities from image $I_a$ to image $I_b$, while $S_{ba}$ corresponds to the lower-left block and encodes similarities in the reverse direction.

For each region pair $(R_i, R_j)$ with $R_i \in \mathcal{R}_a$ and $R_j \in \mathcal{R}_b$, we compute the directional similarity from $R_i$ to $R_j$ as:
\begin{equation}
    s(R_i, R_j) = \frac{1}{|R_i|\,|R_j|} \sum_{p \in R_i} \sum_{q \in R_j} S_{ab}[p, q],
\end{equation}
where $S_{ab}[p, q]$ denotes the similarity from patch $p$ in $I_a$ to patch $q$ in $I_b$.  
Note that the similarity is asymmetric: $s(R_i, R_j)$ is computed from $S_{ab}$, while $s(R_j, R_i)$ is computed from $S_{ba}$.

To determine region-level correspondences, we perform threshold-based greedy matching in both directions.  
Specifically, for each region in $\mathcal{R}_a$, we select all regions in $\mathcal{R}_b$ whose similarity $s(R_i, R_j)$ exceeds a predefined threshold, resulting in the forward match set $\mathcal{M}_{a \rightarrow b}$.  
Likewise, we compute the reverse matches $\mathcal{M}_{b \rightarrow a}$ using $S_{ba}$.  
The final correspondence set is defined as the union $\mathcal{M} = \mathcal{M}_{a \rightarrow b} \cup \mathcal{M}_{b \rightarrow a}$.

\begin{figure*}[t]
\centering
    \includegraphics[width=\linewidth]{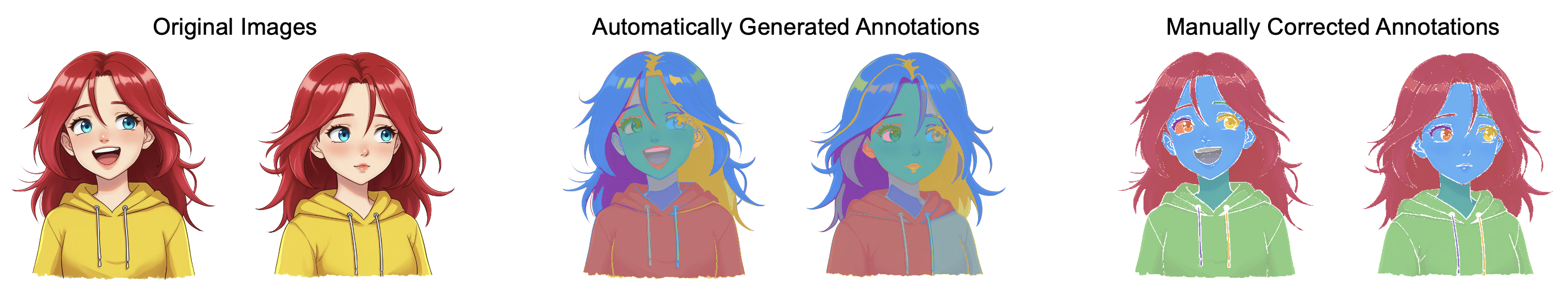}
    \caption{An example of automatic region matching results and manually corrected ground truth. Regions with the same color across the two images indicate a matched region pair, while gray denotes regions without a match.}
  \label{fig:dataset_anno}
\end{figure*}

\section{Annotation Dataset Construction}

To support both training and evaluation of the proposed framework, we require datasets containing region-wise correspondences between pairs of manga line art images.
Since no existing dataset provides such annotations, we construct the necessary data through two separate approaches.

For training, we develop an automatic annotation pipeline that generates large-scale supervision using colored line art as auxiliary input. This process consists of two main stages:
(1) segmenting each line art image into coherent regions based on color guidance;
(2) establishing region-level correspondences between paired line art images using appearance and spatial cues.

For evaluation, however, high-quality ground truth is essential. Therefore, we manually annotate a subset of image pairs with accurate region segmentation and correspondence labels. This manually curated dataset enables reliable benchmarking of our model’s performance under realistic and challenging conditions.

\subsection{Automatically Annotated Training Datasets}
\label{sec:dataset_train}

\noindent\textbf{\textit{{Automatic Intra-Image Segmentation.}}}
Most existing deep learning-based segmentation methods are trained on natural images, as region-segmented datasets for manga line art are scarce.
However, the domain gap between natural images and line art causes these models to struggle with fine-grained segmentation on line art images.
To address this issue, we adopt a rule-based method that leverages the inherent color structure of colored line art. 
We segment colored line art by clustering pixels into $K$ dominant colors, detecting connected regions via color similarity, and merging small fragments based on both color and boundary cues.

\noindent\textbf{\textit{{Automatic Cross-Image Matching.}}}
After obtaining the region segmentation masks, we construct region-wise correspondences between line art pairs. 
We first perform point-to-point matching using LightGlue~\cite{lindenberger2023lightglue}. 
Since the extracted keypoints are often concentrated near structural boundaries and are relatively sparse on line art images, we expand each keypoint to a $5\times5$ pixel neighborhood to increase coverage and improve robustness in subsequent region matching.
For each region in image $I_a$, we filter the matched keypoints by comparing the color of each keypoint’s pixel to the average color of the region. Points with large deviations are discarded. 
We then count the occurrences of target regions in $I_b$ where the remaining keypoints are matched, and assign the region in $I_b$ with the highest vote count as the corresponding match of the region in $I_a$.

Additionally, since point-to-point keypoints tend to be overly concentrated on facial areas of characters, we further apply a coarse matching step for unmatched regions. We adopt a rule-based matching method that considers both positional alignment and color similarity between regions.

\subsection{Manually Refined Evaluation Datasets}
\label{sec:dataset_eval}

While our training pipeline relies on automatically generated region segmentation and correspondences, accurate evaluation requires ground-truth annotations that reflect human-level understanding of semantic regions and their matches.  
To address this need, we construct manually refined evaluation datasets, covering both synthetic and real-world manga line art. These datasets serve as standard benchmarks for assessing model performance.  

As illustrated in Figure~\ref{fig:dataset_anno}, we compare the automatically generated region correspondences with the manually corrected ground truth.  
These comparisons demonstrate that our automatic annotation pipeline provides reasonable supervision for training, while also highlighting the necessity of manual refinement for reliable evaluation.
We will release the full annotation pipeline, manual correction tools, and the evaluation dataset to facilitate future research.

\section{Experiments}

\begin{table*}[t]
\centering
    \setlength\tabcolsep{9pt}
\begin{tabular}{@{}llcccccc@{}}
\toprule
Dataset & Matching Type & AP & Best F1 & Precision & Recall & Top-1 Accuracy & Top-5 Accuracy \\
\midrule
\multirow{3}{*}{ATD} 
& Intra-img ($I_a$) & 88.75 & 79.29 & 74.20 & 85.13 & -- & -- \\
& Intra-img ($I_b$) & 87.36 & 77.17 & 76.65 & 77.71 & -- & -- \\
& Cross-img       & 83.72 & 73.44 & 64.97 & 84.46 & 82.51 & 92.44 \\
\midrule
\multirow{3}{*}{GenAI} 
& Intra-img ($I_a$) & 85.99 & 76.76 & 76.30 & 77.22 & -- & -- \\
& Intra-img ($I_b$) & 88.42 & 79.03 & 79.87 & 78.20 & -- & -- \\
& Cross-img       & 83.49 & 73.39 & 73.73 & 73.05 & 67.73 & 77.42 \\
\bottomrule
\end{tabular}
\caption{
Patch-level evaluation on ATD and GenAI datasets (percentage values).
}
\label{tab:patch}
\end{table*}

\subsection{Datasets for Training and Evaluation}
\label{sec:datasets}

We evaluate our method on both the 3D-rendered PBC dataset~\cite{dai2024learning} and hand-drawn manga-style line art.  
Since our primary focus is correspondence prediction in realistic manga-style line art, we report results on hand-drawn datasets in the main paper and defer PBC results to the supplementary material.

For training, we construct an in-house dataset from short anime clips.  
Frame pairs are sampled every 18 frames, converted to line art using the MangaNinja method~\cite{liu2025manganinja}, and annotated using our automatic region segmentation and matching pipeline introduced in Section~\ref{sec:dataset_train}.  
This produces 200{,}000 image pairs (364{,}015 images), covering a wide range of variations in pose, clothing, and artistic style.

For quantitative evaluation, we construct two benchmark datasets. 
The first is ATD-12K (ATD)~\cite{siyao2021deep}, derived from real anime frames. 
The second is GenAI, a synthetic dataset generated using the Imagen diffusion model~\cite{saharia2022photorealistic}.
We extract line art using the same method as in training and manually refine region correspondences: 25 pairs (50 images) for ATD and 40 pairs (80 images) for GenAI.

\subsection{Baselines}
\label{sec:baseline}

Since no existing method directly tackles our task of region-wise correspondence prediction between unannotated manga line art images, we construct a strong baseline based on existing approaches.
Specifically, we adapt the framework of Dai et al.~\cite{dai2024learning}, originally designed for automatic colorization of pre-segmented line art, and combine it with a high-quality region segmentation method. This design represents one of the strongest reproducible pipelines using publicly available implementations~\cite{feng2025unified}.

In their setting, the model takes a colored reference and an uncolored target line art, both segmented into closed regions, and predicts the colorized target while implicitly learning region correspondences.
To adapt this model to our setting, we first perform rule-based region segmentation on raw line art images. 
For line art $I_a$, each segmented region is assigned a unique color, producing a pseudo-colored version of $I_a$. 
This pseudo-colored $I_a$ and the uncolored $I_b$ are then fed into the model of Dai et al. to obtain a colorized $I_b$. 
Finally, we infer region-wise correspondences between $I_a$ and $I_b$ by matching regions with identical colors.

We implement two rule-based segmentation methods for this baseline:
(1) \textbf{\textit{ClosedRegion}}: a contour-based method that extracts strictly enclosed regions, and  
(2) \textbf{\textit{TrappedBall}}~\cite{zhang2009vectorizing}: an edge-following method that allows small gaps or open contours to remain connected.
Since most regions in hand-drawn manga style line art are not fully closed, we use the TrappedBall method as a more challenging baseline for quantitative evaluation.

\subsection{Implementation Details}
We adopt a ViT-B/16 backbone pre-trained on ImageNet~\cite{dosovitskiy2020image}, and normalize the input images using the standard ImageNet mean and standard deviation to match the requirements of the pre-trained model. 
To fully leverage the capacity of ViT, we use its default patch size of $16 \times 16$.
Our Multiplex Transformer consists of $M = 4$ stacked layers of self-attention and cross-attention.
The model is trained for 20 epochs with a batch size of 16, using the AdamW optimizer and an initial learning rate of $1\times10^{-4}$. A learning rate scheduler combining a warm-up phase with cosine annealing is applied.
The whole process is conducted on NVIDIA A100 GPUs.

\subsection{Quantitative Results}

\noindent\textbf{\textit{{Patch-Level Evaluation.}}}
\label{sec:eval_patch}
We first assess how well our Transformer-based model captures patch-level semantic similarities by comparing the predicted similarity matrix $S$ against the ground-truth correspondence matrix $G$.
We conduct experiments on ATD and GenAI datasets, and report the results in Table~\ref{tab:patch}.

We evaluate both intra-image similarity (within the same image) and cross-image similarity (between image pairs) using the following metrics.
First, we compute precision-recall (PR) curves by thresholding similarity scores and report standard metrics including Average Precision (AP), best F1 score, and the corresponding precision and recall at the optimal threshold.
For cross-image evaluation, we also report top-$K$ accuracy, defined as the percentage of ground-truth matched patches whose true match appears among the top-$K$ most similar patches in the other image.

As shown in Table~\ref{tab:patch}, our model achieves strong intra-image performance on both the ATD and GenAI datasets, with AP scores above 85\% and best F1 scores exceeding 76\%. These results demonstrate the model’s ability to generalize effectively to realistic manga-style line art.  
In the cross-image setting, our model achieves a level of performance comparable to the intra-image case, confirming its ability to learn patch-level correspondences across image pairs. This robustness is largely attributed to our Transformer’s joint learning of intra-image and cross-image features, which enables effective matching even in the absence of color and texture cues.
We also observe that the ATD dataset yields higher top-$K$ accuracy, likely because its domain characteristics are more closely aligned with those of our training data.

\begin{table*}[t]
\centering
    \setlength\tabcolsep{8pt}
\begin{tabular}{@{}lllccccc@{}}
\toprule
Dataset & Method & Type & ARI & mIoU (P$\rightarrow$G) & mIoU (G$\rightarrow$P) & CR & Region Accuracy \\
\midrule
\multirow{6}{*}{ATD}
& \multirow{3}{*}{Baseline}
& Intra-img ($I_a$) & \textbf{64.04} & 13.05 & \textbf{62.11} & 7.23 & -- \\
&  & Intra-img ($I_b$) & \textbf{63.69} & 12.53 & \textbf{61.64} & 7.07 & -- \\
&  & Cross-img & -- & -- & -- & -- & 82.94 \\
\cmidrule(lr){2-8}
& \multirow{3}{*}{Ours}
& Intra-img ($I_a$) & 48.11 & \textbf{31.00} & 33.15 & \textbf{1.41} & -- \\
&  & Intra-img ($I_b$) & 45.06 & \textbf{28.50} & 34.06 & \textbf{1.61} & -- \\
&  & Cross-img & -- & -- & -- & -- & \textbf{84.44} \\
\midrule
\multirow{6}{*}{GenAI}
& \multirow{3}{*}{Baseline}
& Intra-img ($I_a$) & 33.90 & 5.89 & \textbf{44.84} & 13.84 & -- \\
&  & Intra-img ($I_b$) & 34.89 & 5.68 & \textbf{45.01} & 14.22 & -- \\
&  & Cross-img & -- & -- & -- & -- & 72.36 \\
\cmidrule(lr){2-8}
& \multirow{3}{*}{Ours}
& Intra-img ($I_a$) & \textbf{46.23} & \textbf{32.50} & 40.09 & \textbf{1.70} & -- \\
&  & Intra-img ($I_b$) & \textbf{50.92} & \textbf{32.69} & 40.90 & \textbf{1.66} & -- \\
&  & Cross-img & -- & -- & -- & -- & \textbf{78.43} \\
\bottomrule
\end{tabular}
\caption{
Region-level evaluation on ATD and GenAI datasets (percentage values; CR is shown in raw value). Region Precision and Region Recall are computed using predicted region pairs with purity $>$ 0.8.
}
\label{tab:region}
\end{table*}

\noindent\textbf{\textit{{Region-Level Evaluation.}}}
\label{sec:eval_region}
We further evaluate our method at the region level, focusing on two aspects: (1) \textbf{\textit{intra-image region grouping via patch merging}}, and (2) \textbf{\textit{cross-image region correspondence prediction}}. 
Results on the ATD and GenAI datasets are shown in Table~\ref{tab:region}.

To assess the quality of patch clustering within each image, we compare the predicted region groups $\mathcal{R}_a$ and $\mathcal{R}_b$ with the pixel-level ground-truth segmentation using three metrics: Adjusted Rand Index (ARI), mean Intersection over Union from prediction to ground truth (mIoU P$\rightarrow$G), and from ground truth to prediction (mIoU G$\rightarrow$P). 
We also report the Cluster Ratio (CR), which measures the average number of predicted regions per ground-truth region. Ideally, CR should be close to 1, indicating a balanced segmentation. A CR much lower than 1 suggests under-segmentation (i.e., multiple ground-truth regions are merged into one), while a CR significantly higher than 1 implies over-segmentation, where individual regions are unnecessarily fragmented.
For cross-image evaluation, we adopt a strategy inspired by visual relationship detection~\cite{xu2017scene}. Specifically, we compute region-wise precision and recall based on region pairs whose predicted correspondences have a purity score greater than 0.8. This ensures that only meaningful and consistent matches are considered.

In our experiments, we observe that maintaining CR within the range of 1.4 to 1.7 generally leads to visually pleasing and semantically consistent region segmentation results. Therefore, we fix CR within this range when searching for the optimal balance between region-wise precision and recall. 
We use the baseline method introduced in Section~\ref{sec:baseline}, where region segmentation is performed using the TrappedBall method. 

\begin{figure*}[t]
\centering
    \subfloat[Intra-Image Region Segmentation]{
        \includegraphics[width=0.7\textwidth]{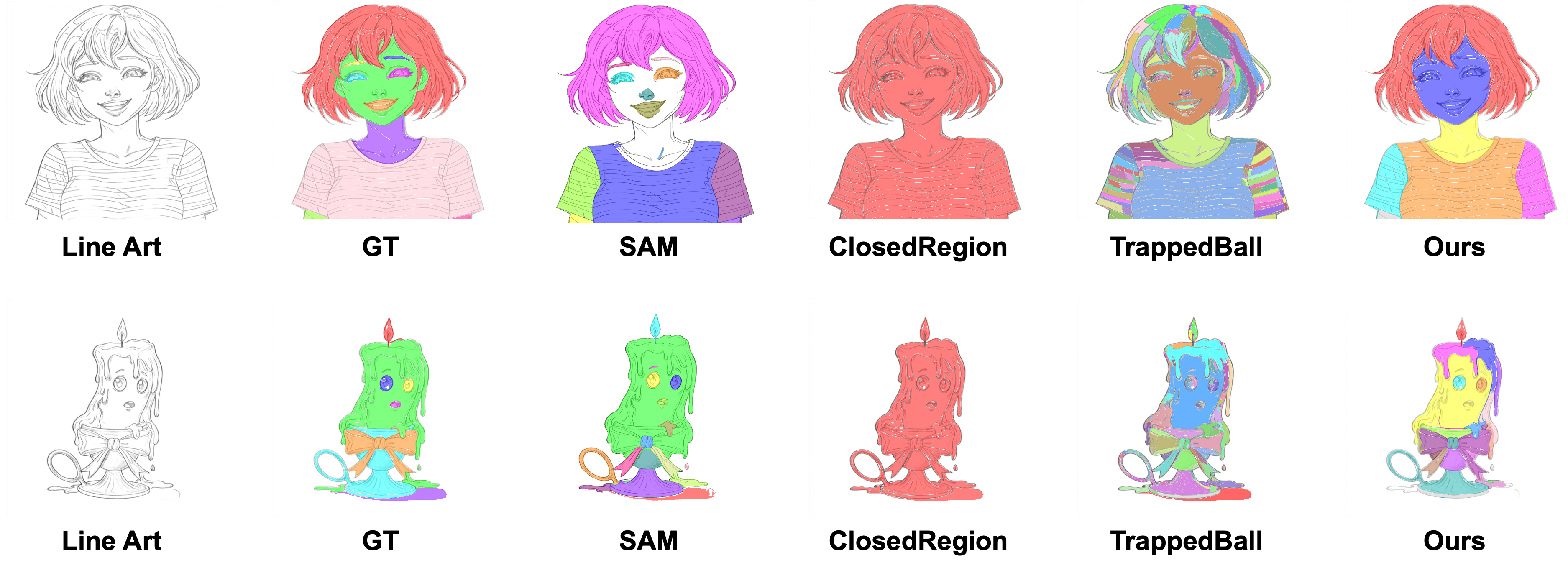}
    }\\[4mm]
    \subfloat[Cross-Image Region Matching]{
        \includegraphics[width=0.98\textwidth]{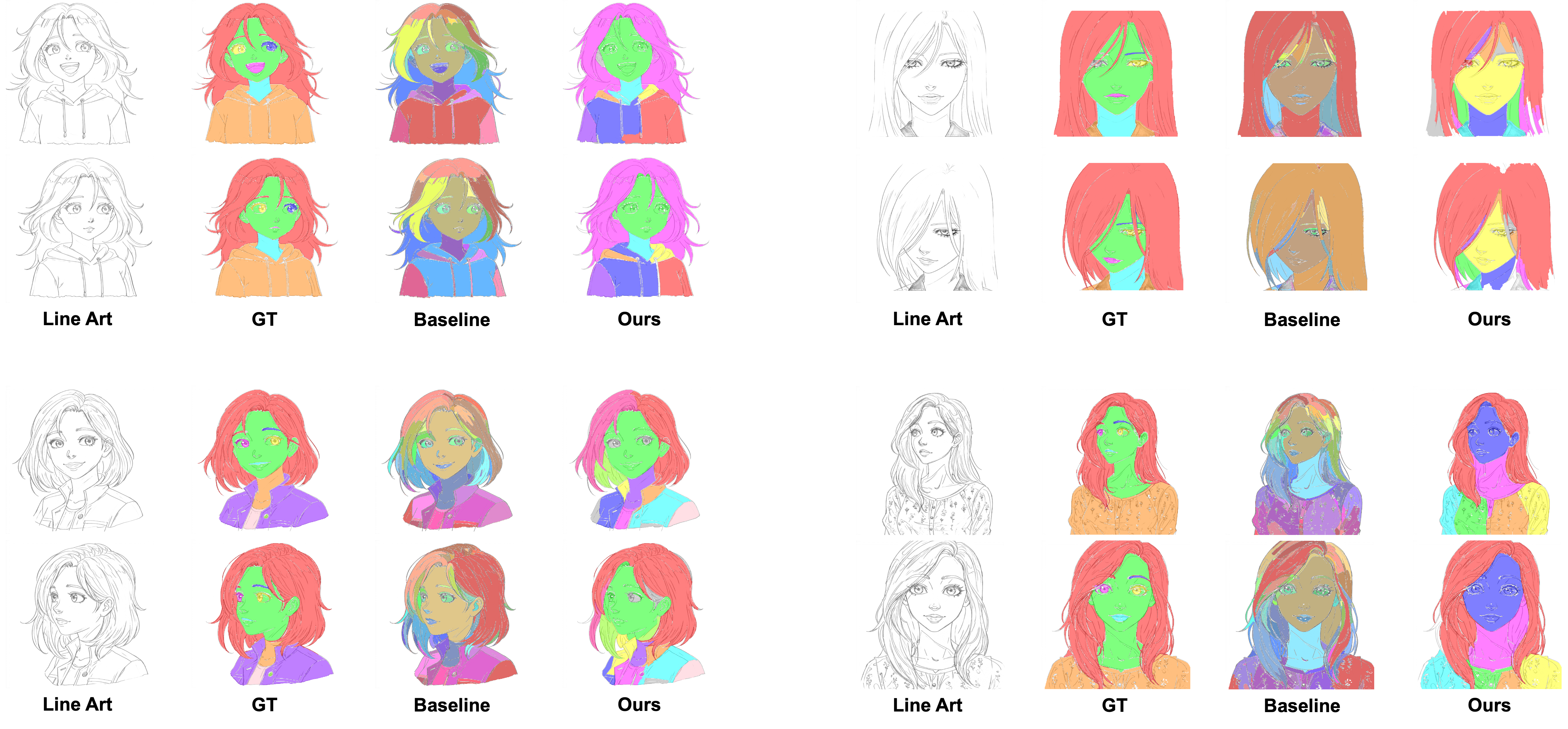}
    }
    \caption{
    Visualization examples on the GenAI dataset. In the region segmentation results, the same color within an image denotes a single region, and white indicates the background. 
    In the region matching results, the same color across two images indicates a matched region pair, while gray denotes regions without a match.
    }
    \label{fig:vis}
\end{figure*}

As shown in Table~\ref{tab:region}, the baseline presents a challenging reference point but exhibits a clear tendency toward over-segmentation. This is evidenced by its smaller mIoU (P$\rightarrow$G), larger mIoU (G$\rightarrow$P), and a Cluster Ratio significantly greater than 1.
This tendency can also be observed qualitatively in Figure~\ref{fig:vis}~(a), where \textbf{TrappedBall} corresponds to the segmentation results of the baseline method.

For cross-image region matching, our method achieves higher precision than the baseline, with a more noticeable improvement on the GenAI dataset. This is because the ATD dataset consists of frame pairs sampled from adjacent animation frames, where character motion and pose variations are relatively small, whereas the GenAI dataset presents a more challenging scenario with larger differences in structure and appearance.
However, region recall remains relatively low (30--35\%), suggesting that our method still struggles to match small fragmented regions or large semantic regions that are inconsistently segmented (e.g., hair or clothing divided into multiple parts). 
This highlights the inherent difficulty of consistent region matching in highly abstract line art.

\begin{figure}[t]
\centering
    \subfloat[Intra-Image Matching]{
       \includegraphics[width=.48\linewidth]{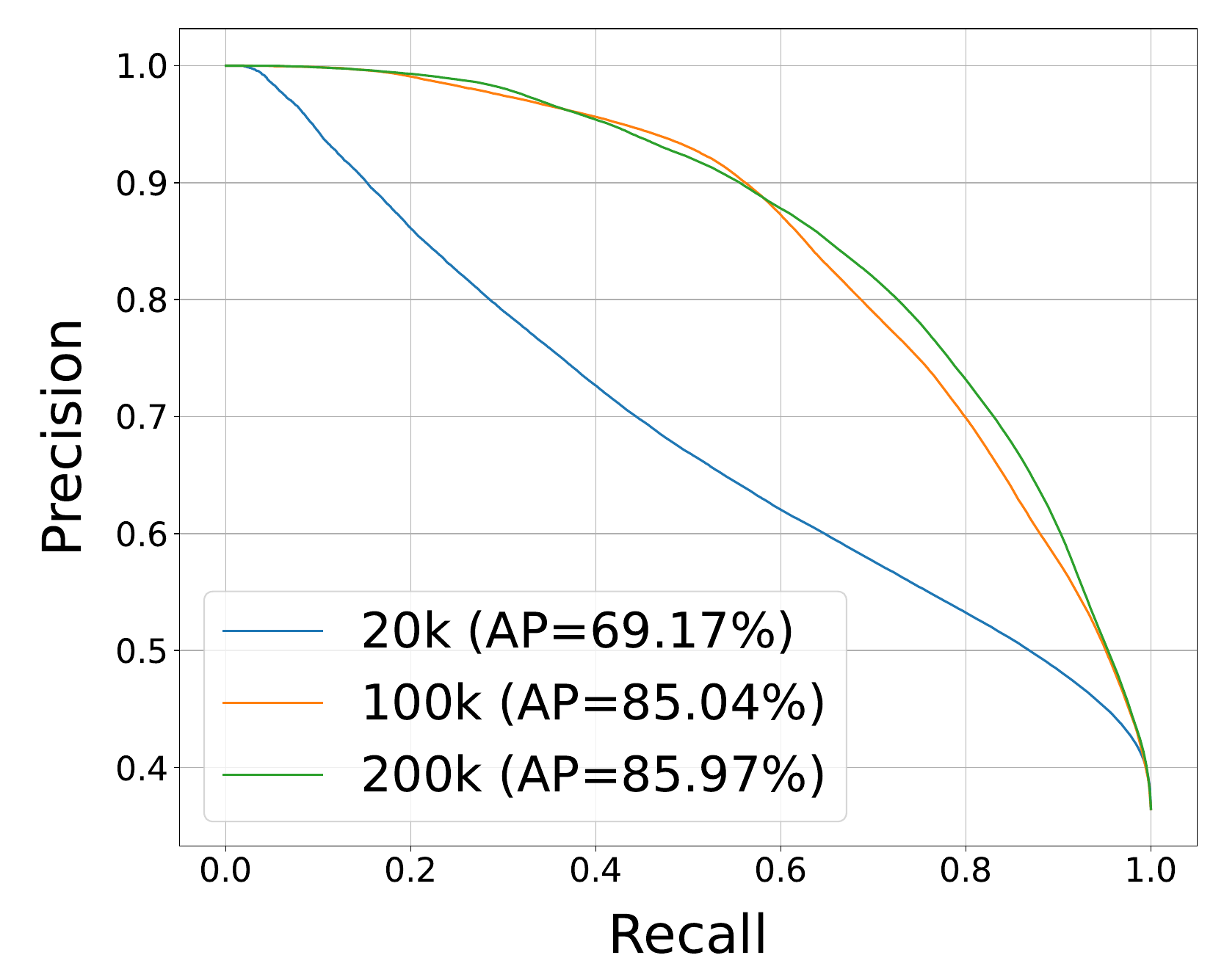}}
    \label{a}
    \subfloat[Cross-Image Matching]{
        \includegraphics[width=.48\linewidth]{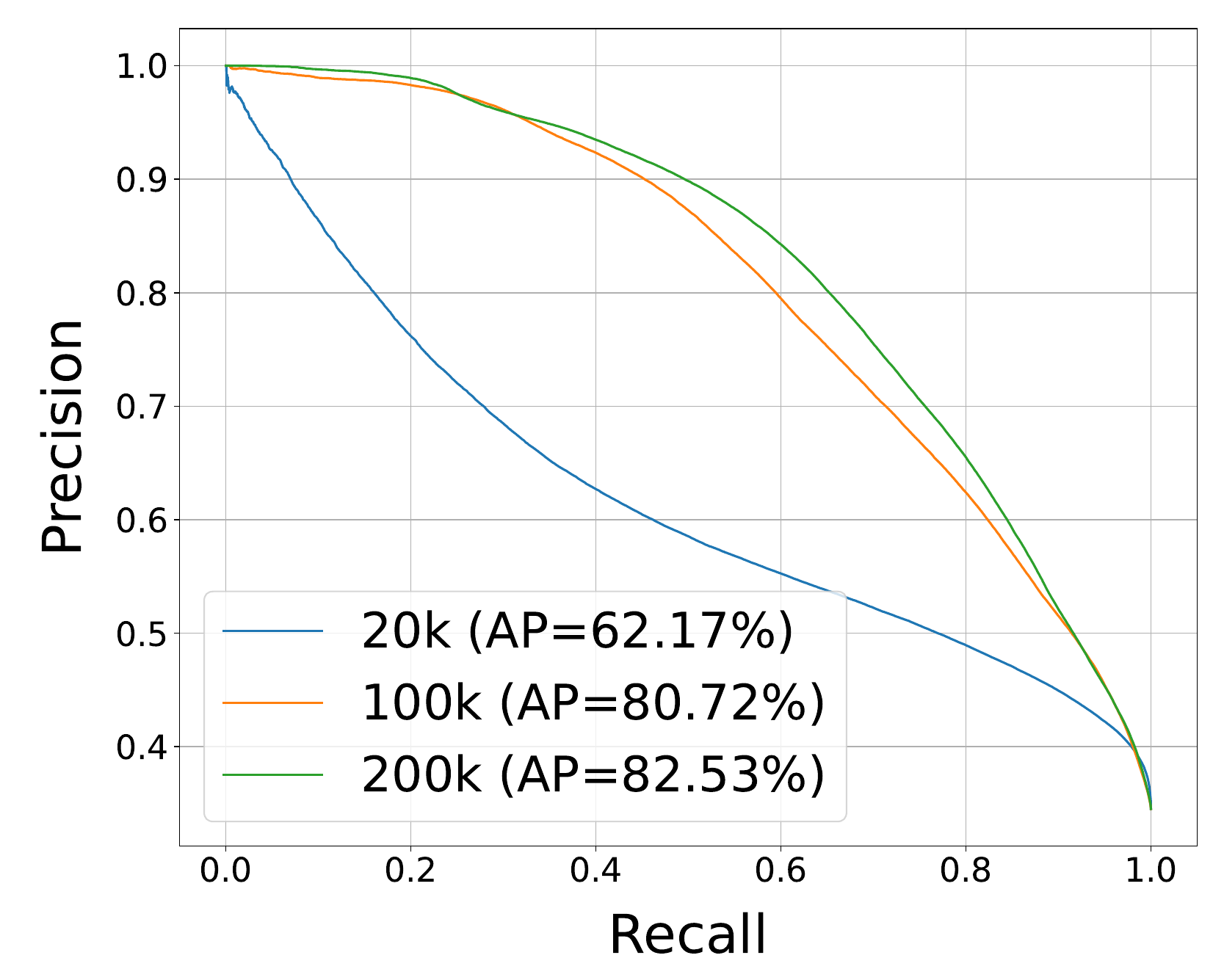}}
    \label{b}
  \caption{Patch-level Precision–Recall (PR) curves on the GenAI dataset under different training set scales.}
  \label{fig:ablation_pr_curve}
\end{figure}

\noindent\textbf{\textit{{{Ablation Study.}}}}
\label{sec:ablation}
We evaluate how the scale of the training dataset affects model performance.  
Patch-level Precision-Recall (PR) curves on the GenAI dataset are shown in Figure~\ref{fig:ablation_pr_curve}.  
As the size of the training set increases, both intra-image and cross-image matching consistently improve, indicating that larger datasets enable the model to learn more stable structural representations.  

\subsection{Qualitative Results}
\label{sec:eval_vis}

We qualitatively evaluate our proposed method on the GenAI dataset, with representative examples shown in Figure~\ref{fig:vis}.  
As in our quantitative evaluation, we analyze the results from two aspects: (1) \textbf{\textit{intra-image region segmentation}} and (2) \textbf{\textit{cross-image region matching}}.

For region segmentation, we compare our method with the two rule-based methods introduced in Section~\ref{sec:baseline} (ClosedRegion and TrappedBall).  
In addition, we include the Segment Anything Model (SAM)~\cite{kirillov2023segment}, a strong baseline for natural images.  
As shown in Figure~\ref{fig:vis}~(a), the scarcity of closed contours in hand-drawn manga-style line art causes the ClosedRegion method to fail to produce meaningful segmentation.  
SAM, on the other hand, often merges distinct semantic parts (e.g., the face) with the background and cannot generate fine-grained boundaries due to the absence of color, texture, and shading cues.  
The TrappedBall method yields finer segmentation but, as shown in Table~\ref{tab:region}, tends to over-segment the image.  
In contrast, our method produces more semantically coherent regions, such as grouping facial features, hair, and clothing into consistent and meaningful structural units.

For region matching, our approach produces more accurate and perceptually consistent correspondences compared with the baseline methods, as illustrated in Figure~\ref{fig:vis}~(b).
Nevertheless, as discussed in Section~\ref{sec:eval_region}, there remains a gap compared with manually annotated ground truth for cross-image region matching.
For instance, to avoid over-segmentation, we merge small regions during post-processing, which can lead to the loss of fine-grained semantic details -- such as eyes and eyebrows being merged into larger facial regions.  
We acknowledge this trade-off as a current limitation and plan to address it through improvements to both the model architecture and the post-processing pipeline.


\section{Conclusion}
\label{sec:conclusion}

In this paper, we explored a realistic and previously underexplored task: predicting region-wise correspondence between manga line art images without any prior segmentation or annotations.  
We proposed a Transformer-based framework that jointly learns intra-image and cross-image features from raw line art images, and introduced a post-processing method that aggregates and refines the patch-level predictions into region-level correspondences.  
To support training and evaluation, we developed an automatic annotation pipeline and manually refined a subset of the annotations to construct benchmark evaluation datasets.  
Experiments on multiple datasets demonstrate our model achieves high patch-level accuracy and produces coherent region-level correspondences.
In future work, we aim to extend our approach to handle more diverse manga styles and complex scenes, and to improve its robustness in real-world scenarios.  
We also plan to apply our region correspondence model to downstream tasks such as automatic line art colorization and frame interpolation, where consistent region-wise understanding plays a critical role.

{
    \small
    \bibliographystyle{ieeenat_fullname}
    \bibliography{main}
}

\clearpage
\setcounter{page}{1}
\maketitlesupplementary

\newcommand\beginsupplement{%
        \setcounter{table}{0}
        \renewcommand{\thetable}{\Alph{table}}%
        \setcounter{figure}{0}
        \renewcommand{\thefigure}{\Alph{figure}}%
     }
\beginsupplement
\appendix

\section{Ablation Study}

This section presents the complete quantitative results for the training-scale ablation introduced in Section~\ref{sec:ablation} of the main paper.  
Patch-level and region-level performance on the GenAI dataset is reported in Table~\ref{tab:ablation_patch} and Table~\ref{tab:ablation_region}, respectively.
As shown in these tables, scaling up the training data consistently improves structural correspondence, leading to better intra-image region grouping and cross-image region matching.

\begin{table*}[t]
\centering
    \setlength\tabcolsep{12pt}
\begin{tabular}{@{}llcccccc@{}}
\toprule
Training Pairs & Matching Type & AP & Best F1 & Precision & Recall & Top-1 Accuracy \\
\midrule
\multirow{3}{*}{20k} 
& Intra-img ($I_a$) & 69.11 & 64.37 & 53.23 & \textbf{81.42} & -- \\
& Intra-img ($I_b$) & 69.23 & 63.57 & 53.38 & \textbf{78.56} & -- \\
& Cross-img       & 62.17 & 60.74 & 48.84 & \textbf{80.31} & 51.58 \\
\midrule
\multirow{3}{*}{100k} 
& Intra-img ($I_a$) & 84.82 & 75.70 & \textbf{75.90} & 75.51 & -- \\
& Intra-img ($I_b$) & 85.26 & 74.34 & 71.35 & 77.59 & -- \\
& Cross-img       & 80.72 & 70.73 & 68.92 & 72.65 & \textbf{68.46} \\
\midrule
\multirow{3}{*}{200k} 
& Intra-img ($I_a$) & \textbf{85.32} & \textbf{76.60} & 74.69 & 78.60 & -- \\
& Intra-img ($I_b$) & \textbf{86.62} & \textbf{76.62} & \textbf{77.47} & 75.80 & -- \\
& Cross-img       & \textbf{82.53} & \textbf{72.79} & \textbf{72.96} & 72.63 & 66.86 \\
\bottomrule
\end{tabular}
\caption{Patch-level evaluation on the GenAI dataset under different training set scale.}
\label{tab:ablation_patch}
\end{table*}

\begin{table*}[t]
\centering
    \setlength\tabcolsep{9pt}
\begin{tabular}{@{}llcccccc@{}}
\toprule
Training Pairs & Matching Type & ARI & mIoU (P→G) & mIoU (G→P) & CR & Region Accuracy \\
\midrule
\multirow{3}{*}{20k}
& Intra-img ($I_a$) & 31.32 & 26.97 & 35.22 & 1.94 & -- \\
& Intra-img ($I_b$) & 34.83 & 28.50 & 35.87 & 1.90 & -- \\
& Cross-img       & -- & -- & -- & -- & 65.72 \\
\midrule
\multirow{3}{*}{100k}
& Intra-img ($I_a$) & 49.53 & 35.56 & \textbf{38.58} & 1.37 & -- \\
& Intra-img ($I_b$) & 54.51 & 36.93 & \textbf{40.76} & 1.36 & -- \\
& Cross-img       & -- & -- & -- & -- & 66.20 \\
\midrule
\multirow{3}{*}{200k}
& Intra-img ($I_a$) & \textbf{50.21} & \textbf{38.25} & 38.29 & \textbf{1.26} & -- \\
& Intra-img ($I_b$) & \textbf{55.33} & \textbf{39.41} & 39.47 & \textbf{1.20} & -- \\
& Cross-img       & -- & -- & -- & -- & \textbf{70.44} \\
\bottomrule
\end{tabular}
\caption{
Region-level evaluation on the GenAI dataset under different training set scales.
}
\label{tab:ablation_region}
\end{table*}

\section{Experiments on 3D-Rendered Line Art}

As noted in Section~\ref{sec:datasets} of the main paper, in addition to experiments on hand-drawn manga-style line art, we also conduct training and evaluation on the 3D-rendered PaintBucket-Character (PBC) dataset~\cite{dai2024learning}. This section provides details of the dataset conduction, and shows quantitative and qualitative results on the PBC dataset.

\subsection{Dataset}

The PBC dataset is created by extracting contour lines from 3D-rendered anime characters.  
This rendering-to-line-art process produces clean and fully closed contours with clearly separable semantic parts, making region segmentation substantially easier than in hand-drawn line art.

\begin{figure}[t]
\centering
    \includegraphics[width=\linewidth]{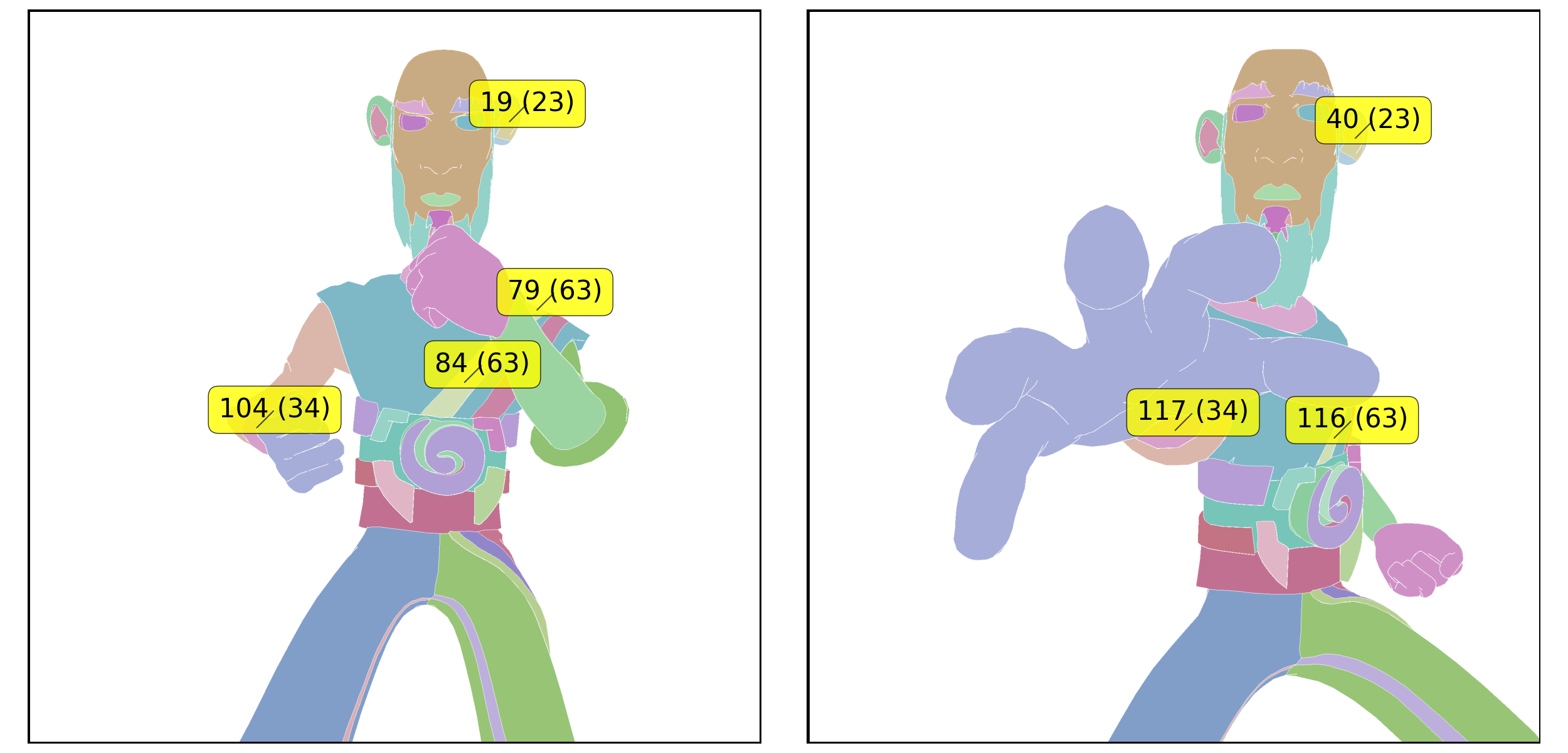}
    \caption{Example of region-level correspondence annotations in the PBC dataset. Matched regions are visualized using the same color across two images. Yellow boxes display the region ID and label ID of each region in the format \texttt{Region ID (Label ID)}.}
  \label{fig:pbc_anno}
\end{figure}

The dataset contains over 10{,}000 synthetic line art images paired with part-level semantic labels under consistent character poses and viewpoints.  
We sample line art pairs by selecting frames separated by seven frames within the same animation sequence, and define region correspondences based on shared semantic part labels.
For example, region \texttt{19} in the left image of Figure~\ref{fig:pbc_anno} and region \texttt{40} in the right image share label ID \texttt{23}, indicating a semantic correspondence.

In our experiments, we randomly sample 300 image pairs (595 images) for testing, and use the remaining 10,961 pairs (11,336 images) for training.

\subsection{Quantitative Results}

Patch-level and region-level evaluation results on the PBC dataset are reported in Table~\ref{tab:x_patch} and Table~\ref{tab:x_region}, respectively.

As shown in Table~\ref{tab:x_patch}, our model achieves near-perfect intra-image patch matching performance, with AP scores above 99.9\% and best F1 scores exceeding 98.5\% for both $I_a$ and $I_b$. 
This demonstrates that the model effectively captures internal region consistency in well-structured synthetic line art.

At the region level, we tune post-processing parameters to preserve visually coherent segmentation while maximizing prediction accuracy. 
As shown in Table~\ref{tab:x_region}, our method performs well on both region segmentation and region correspondence, achieving a correspondence accuracy of 59.9\%. 
However, because the PBC dataset contains many small and fine-grained regions, the Cluster Ratio (CR) and region recall remain relatively low, typically around 30\%. 
While reducing over-segmentation and improving region recall represent a fundamental trade-off on this dataset, achieving a better balance will require a deeper structural understanding of abstract line art, highlighting an important direction for future work.

\begin{table*}[t]
\centering
    \setlength\tabcolsep{9pt}
\begin{tabular}{@{}lcccccc@{}}
\toprule
Matching Type & AP & Best F1 & Precision & Recall & Top-1 Acc. & Top-5 Acc. \\
\midrule
Intra-img ($I_a$) & 99.92 & 98.57 & 99.06 & 98.08 & --     & -- \\
Intra-img ($I_b$) & 99.93 & 98.66 & 98.92 & 98.40 & --     & -- \\
Cross-img       & 98.86 & 95.23 & 96.34 & 94.15 & 85.44 & 91.35 \\
\bottomrule
\end{tabular}
\caption{
Patch-level evaluation on PBC datasets (percentage values).
}
\label{tab:x_patch}
\end{table*}

\begin{table*}[t]
\centering
    \setlength\tabcolsep{9pt}
\begin{tabular}{@{}lccccc@{}}
\toprule
Matching Type & ARI & mIoU (P$\rightarrow$G) & mIoU (G$\rightarrow$P) & CR & Region Accuracy \\
\midrule
Intra-img ($I_a$) & 83.05 & 60.20 & 20.30 & 0.32 & -- \\
Intra-img ($I_b$) & 83.86 & 60.34 & 19.97 & 0.32 & -- \\
Cross-img       & --    & --    & --    & --     & 59.92 \\
\bottomrule
\end{tabular}
\caption{
Region-level evaluation on PBC datasets (percentage values; CR is shown in raw value). Region Precision and Region Recall are computed using predicted region pairs with purity $>$ 0.8.
}
\label{tab:x_region}
\end{table*}

\subsection{Qualitative Results}

\begin{figure*}[t]
\centering
    \includegraphics[width=\linewidth]{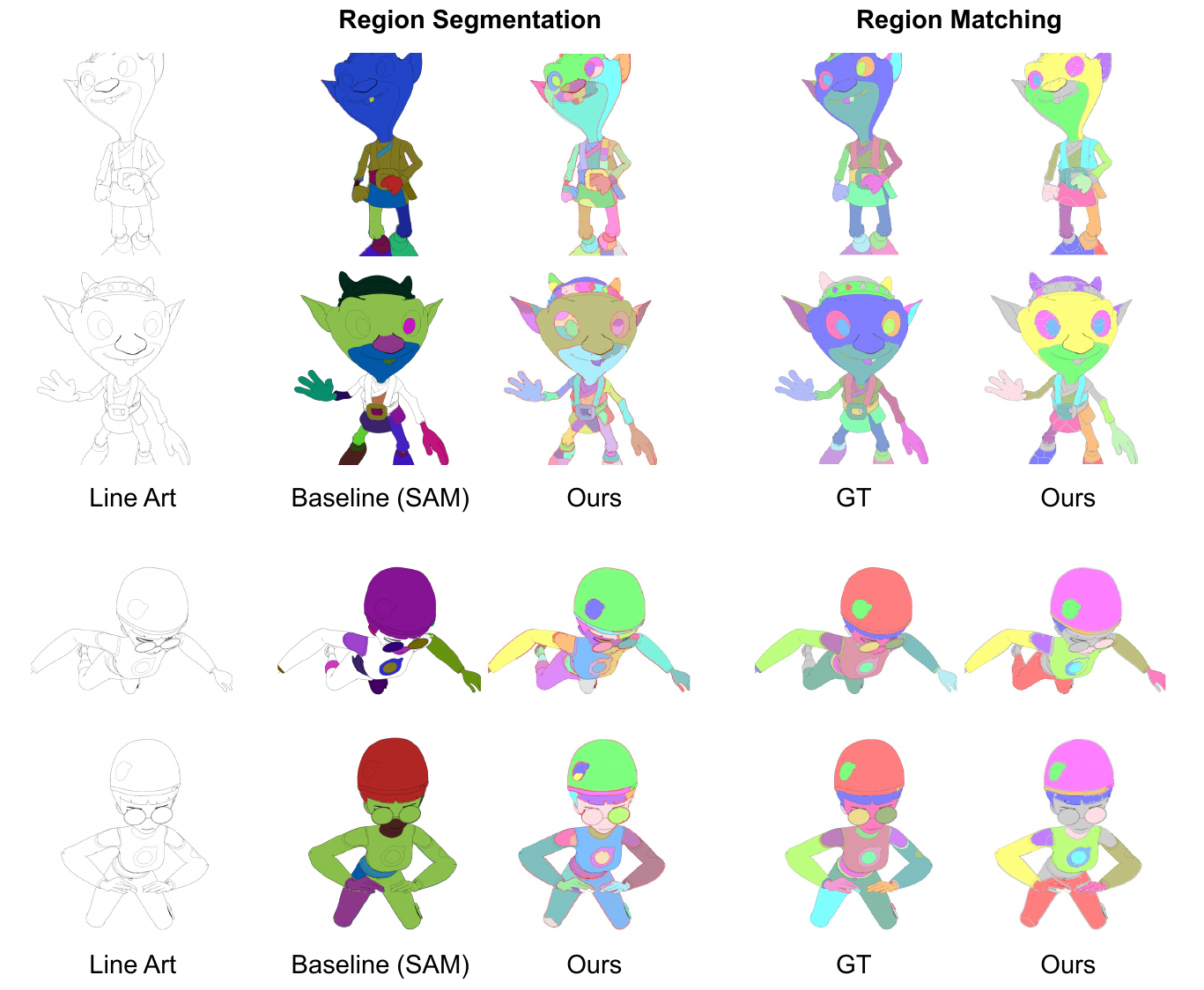}
    \caption{Visualization examples on the PBC dataset. In the region segmentation results, the same color within an image denotes a single region, and white indicates the background. 
    In the region matching results, the same color across two images indicates a matched region pair, while gray denotes regions without a match.}
  \label{fig:vis_pbc}
\end{figure*}

Following the qualitative analysis in Section~\ref{sec:eval_vis} of the main paper, we evaluate our method on the PBC dataset from two aspects: (1) \textbf{\textit{intra-image region segmentation}} and (2) \textbf{\textit{cross-image region matching}}. Representative examples are shown in Figure~\ref{fig:vis_pbc}.

For region segmentation, we compare our method with the Segment Anything Model (SAM)~\cite{kirillov2023segment}.  
As observed in Figure~\ref{fig:vis_pbc}, SAM often merges distinct semantic parts (e.g., face or clothing) with the background and fails to produce fine-grained boundaries due to the lack of color, texture, and shading cues in line art.  
In contrast, our method produces more precise and semantically coherent region groups.

For region matching, our approach accurately aligns major semantic regions, including cases with noticeable pose variation or shape deformation.  
The resulting region correspondences generally agree with human perception and demonstrate the applicability of our framework to clean, closed-contour line art.

\end{document}